\documentclass[11pt]{article}

\usepackage[preprint]{acl}

\usepackage{times}
\usepackage{latexsym}

\usepackage[T1]{fontenc}

\usepackage[utf8]{inputenc}

\usepackage{microtype}

\usepackage{inconsolata}

\usepackage{graphicx}

\title{Beyond MedQA: Towards Real-world Clinical Decision Making \\in the Era of LLMs}

\author{
 \textbf{Yunpeng Xiao\textsuperscript{1}},
 \textbf{Carl Yang\textsuperscript{1}},
 \textbf{Mark Mai\textsuperscript{2}},
 \textbf{Xiao Hu\textsuperscript{3}},
 \textbf{Kai Shu\textsuperscript{1}}
\\
 \textsuperscript{1}Department of Computer Science, Emory University,
\\
 \textsuperscript{2}Children's Healthcare of Atlanta,
\\
 \textsuperscript{3}Nell Hodgson Woodruff School of Nursing, Emory University
\\
 \fontfamily{qcr}\selectfont{
   \{yunpeng.xiao, j.carlyang, xiao.hu, kai.shu\}@emory.edu
 }
\\
\fontfamily{qcr}\selectfont{mark.mai@choa.org}
}

\begin{document}
\maketitle
\begin{abstract}
Large language models (LLMs) show promise for clinical use. They are often evaluated using datasets such as MedQA. However, Many medical datasets, such as MedQA, rely on simplified Question-Answering (Q\&A) that under represents real-world clinical decision-making. Based on this, we propose a unifying paradigm that characterizes clinical decision-making tasks along two dimensions: Clinical Backgrounds and Clinical Questions. As the background and questions approach the real clinical environment, the difficulty increases. We summarize the settings of existing datasets and benchmarks along two dimensions. Then we review methods to address clinical decision-making, including training-time and test-time techniques, and summarize when they help. Next, we extend evaluation beyond accuracy to include efficiency, explainability. Finally, we highlight open challenges. Our paradigm clarifies assumptions, standardizes comparisons, and guides the development of clinically meaningful LLMs.
\end{abstract}

\section{Introduction}

In the fast-evolving field of artificial intelligence, large language models (LLMs) have made remarkable progress, showcasing impressive capabilities across diverse applications and opening new possibilities for innovation \cite{hurst2024gpt, achiam2023gpt, roumeliotis2023chatgpt, liu2024deepseek, guo2025deepseek}. The integration of large foundational models into real-world medical practice holds tremendous promise, offering the potential to transform clinical workflows and patient care \cite{singhal2023large, singhal2025toward, wu2024medjourney}. Nowadays, more and more medical fine-tuned large language models are being developed, enriching the practice of professionals and researchers \cite{chen2023meditron, saab2024capabilities, wang2023huatuo, wu2024pmc, sellergren2025medgemma}. 

LLM as a tool needs to be tested before implementation. One aspect of interest to developers of LLM is the performance. Towards testing performance, the medical field has been focused on using datasets, most of which were sourced from standard tests. Among these datasets, the most common are question-answering datasets, including MedQA \cite{jin2021disease}, MMLU \cite{hendrycks2020measuring}, MedmcQA \cite{pal2022medmcqa}, MedxpertQA \cite{zuo2025medxpertqa}, Pubmed QA \cite{jin2019pubmedqa}, and so on. Each case typically contains context (patient history, laboratory data, etc.), a question, and multiple answer options. The model needs to choose the most appropriate option. These datasets have many advantages, such as standardization and easy evaluation of indicators, and have therefore been widely used to evaluate the performance of large language models in the medical field \cite{sellergren2025medgemma, saab2024capabilities, xie2024me}.

However, there are still many differences between the LLM-powered clinical use cases and the question-and-answer (Q\&A). For example, during the patient’s first clinical encounter, the doctor may not be able to obtain the patient's medical history, test results, and other information in advance; he or she must ask about the patient's condition and ask for necessary laboratory tests while avoiding meaningless questions and laboratory tests. But medical Q\&A can often provide this information directly and precisely. The process of making choices in a clinical environment is called clinical decision making, which is a multistep process that requires gathering and synthesizing data from diverse sources and continuously evaluating the facts to reach an evidence-based decision on a patient’s diagnosis and treatment \cite{hager2024evaluation}. Therefore, medical Q\&A can be understood as a clinical decision-making process that simplifies the data collection and analysis process and simplifies clinical solutions. Such simplification promotes the standardization of large language model evaluation in the medical field, but it also raises questions about whether large language models can be applied in actual clinical scenarios.

Some existing research has begun to recognize the gap between medical Q\&A and real-world clinical decision-making, and has begun to explore more realistic task settings. For example, MedQA can be transformed into a doctor-patient dialogue, where doctors must ask patients questions or request tests to gradually obtain information before ultimately making a decision. However, the settings for these tasks vary. For example, in addition to converting MedQA to a doctor-patient dialogue \cite{schmidgall2024agentclinic, li2024mediq, li2025aligning}, recent work directly adds irrelevant content or modifies options in MedQA \cite{pan2025beyond}. Whether these settings truly reflect the clinical decision-making capabilities of large models, and to what extent, is not addressed by the unified paradigm. 

For the reasons mentioned above, this survey aims to propose a paradigm for clinical decision-making tasks. Based on existing works, we evaluate the complexity of a clinical medical decision-making task from two dimensions: Clinical Background and Clinical Questions. 

We summarize our contributions as follows:
\begin{itemize}
    \item We define the clinical decision-making from the perspective of task setting, which provides a reference for other researchers in the future.
    \item We propose a paradigm to describe the real-world clinical decision-making tasks and summarize the existing literature.
    \item We propose several open issues for future clinical decision-making tasks.
\end{itemize}


\section{Definition of Clinical Decision-Making}
Previous definitions of clinical decision-making were summarized in the clinical context, or healthcare \cite{banning2008review,thompson2002clinical,hager2024evaluation}. In the era of large language models, it is important to address the definition of clinical decision-making because it could facilitate the standardization of LLMs development, research, and evaluation in clinical decision-making. According to our observation and existing literature, different medical datasets and clinical decision-making tasks can be divided into two parts:

\textbf{Clinical Backgrounds}. Before making a decision, doctors need to collect necessary information, such as clinical conditions and patient history, which is collectively called the clinical background. Different background settings exist in different tasks. We divide them into four categories: (1) No Background; (2) Precise Information Background; (3) Rich Information Background; (4) Incomplete Information Background. 


\begin{figure*}[!t]
\centering
\includegraphics[width=2\columnwidth]{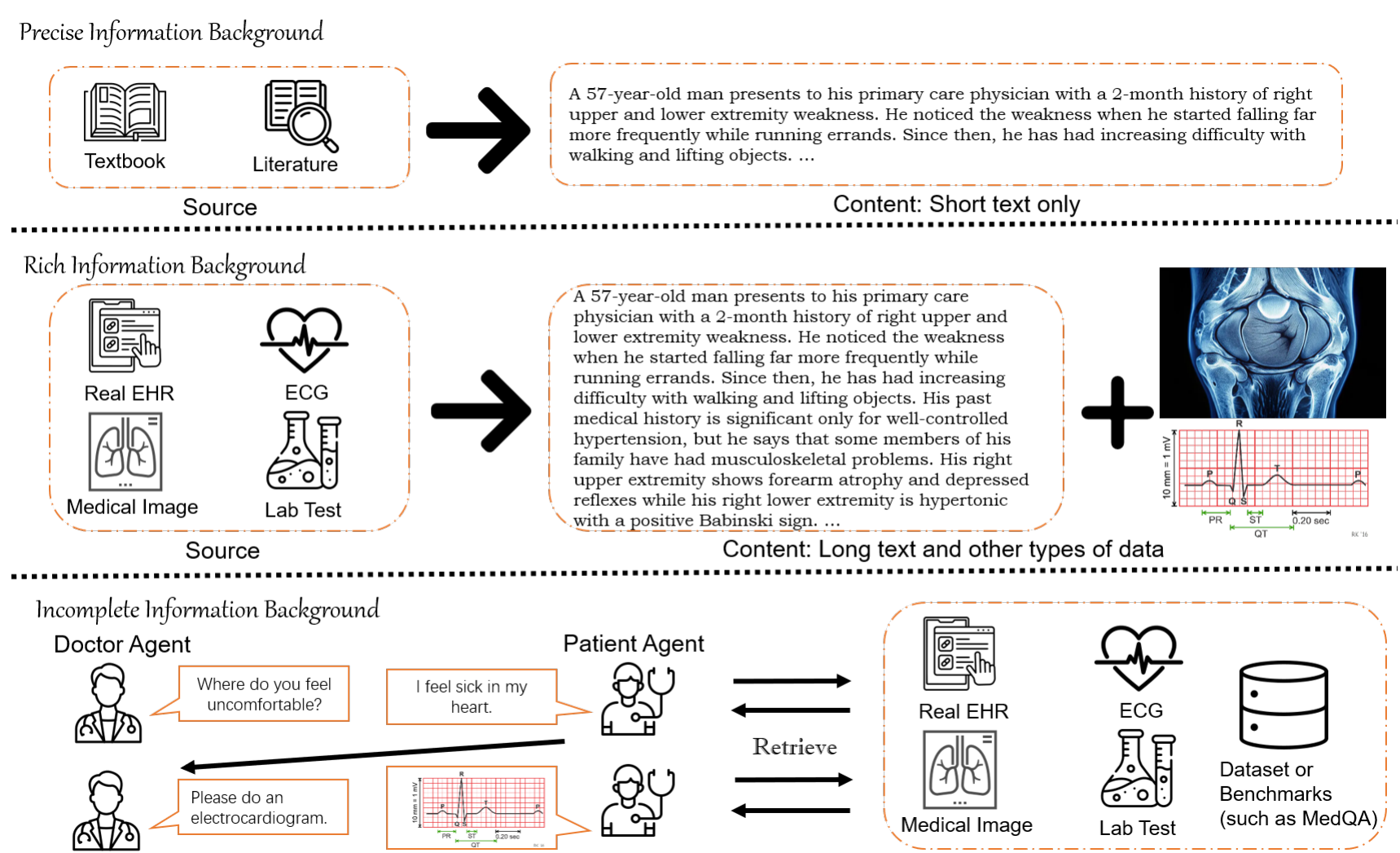} 
\caption{Three different clinical settings (No clinical background setting is not shown in the figure).}
\label{basic_factors}
\end{figure*}

\textbf{Clinical Questions}. In most cases, clinical questions represent decisions that need to be made in a clinical background. In many research works and surveys, “tasks” and “questions” are often used synonymously, and “questions” are categorized according to specific medical tasks, such as EHR summary questions, disease diagnosis questions, etc. In this paper, we classify questions based on the question types. We divide them into four categories: (1) True or false questions, (2) Multiple choice questions, (3) Short-answer questions, and (4) Open-ended questions. 


\begin{figure*}[!t]
\centering
\includegraphics[width=2\columnwidth]{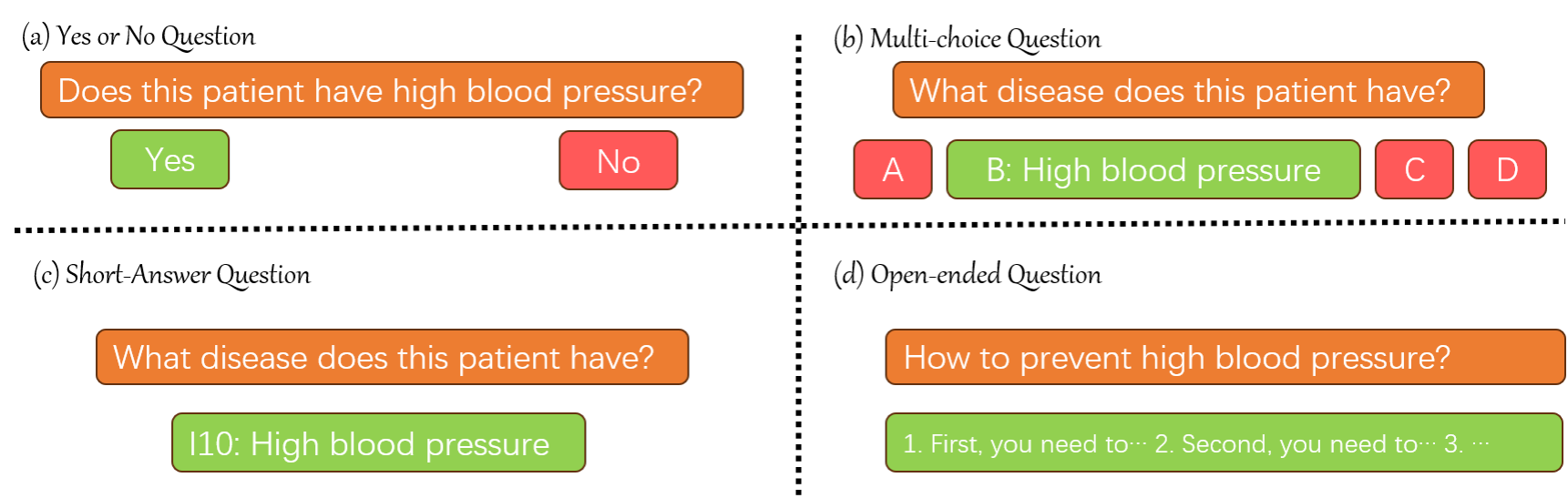} 
\caption{Four different clinical question settings.}
\label{basic_factors}
\end{figure*}

We show different backgrounds and question settings in Figures 1 and 2. The combination of background and questions constitutes the case for clinical decision-making. In clinical practice, rich information background and incomplete information background, especially incomplete information background, and the short-answer questions and open-ended questions, especially open-ended questions, are common. However, most current medical question-answering datasets still use precise information backgrounds and multiple-choice questions. We will introduce each background and question in detail in the next two sections.

\section{Clinical Backgrounds}
Clinical backgrounds are part of clinical decision-making. In this section, we introduce various background settings for clinical decision-making. These tasks have different settings; they can be the original datasets or benchmarks formed by adding various constraints to the original datasets. 

\subsection{No Clinical Backgrounds}
This type of task often only includes a medical question without any context, such as "What are the symptoms of diabetes?" These simple knowledge questions appear in many medical question-answering datasets and benchmarks \cite{zhang2025llmeval, jin2019pubmedqa}. They just test LLMs' ability to master medical knowledge. However, they often appear together with cases with a precise information background in medical question answering datasets \cite{liu2023benchmarking, jin2021disease}.

\subsection{Precise Information Background}
This is the most commonly used task setting in most datasets or benchmarks for clinical decision-making tasks. Some datasets use cases from textbooks \cite{jin2021disease, hendrycks2020measuring}, while others use peer-reviewed literature \cite{jin2019pubmedqa, lin2023pmc}. What they have in common is that, as teaching or published texts, most clinical environments have been simplified, and the resulting background information is dense, so the correct answers can be obtained by using existing medical knowledge and simple reasoning. This type of data set is often plain text data \cite{jin2025medethiceval, pal2022medmcqa, jin2021disease, hendrycks2020measuring, liu2023benchmarking}. Some work has also begun to explore using LLM to generate specific backgrounds \cite{jin2025medethiceval}. Specifically, researchers use textbooks or guidelines as prompts and ask LLMs to generate a series of clinical backgrounds. 

\subsection{Rich Information Background}

Compared to precise background information, this type of background information is richer; it often contains multimodal data; however, as the data increases, it also introduces redundant information and noise. At the same time, they put forward higher requirements on the retrieval and reasoning capabilities of LLMs; one example is that the texts in these tasks are usually longer, and long texts can lead to a decline in the reasoning ability of LLMs \cite{liu2025comprehensive,adams2025longhealth}.

Some datasets and benchmarks are natural extensions of the precise information background. Their sources are still textbooks or medical literature, but compared to the precise information background, the texts are often longer \cite{adams2025longhealth} and include data from other modalities, such as medical images and electrocardiograms, forming multi-modal data backgrounds \cite{lin2023pmc, zuo2025medxpertqa, ben2019vqa, singhal2025toward, lau2018dataset, xie2024medtrinity, zhang2025medtvt}. In addition to using real images, some work has also begun to explore synthetic medical images \cite{zhaobenchmark}. 

Compared to datasets generated directly from textbooks and literature, datasets constructed directly from clinical data, such as the MIMIC dataset \cite{johnson2016mimic,johnson2023mimic}, contain a lot of information about patients. For a specific clinical question, only some of the information may be useful, while other information may be redundant. For example, in the task of predicting the length of hospital stay and mortality rate of COVID-19 patients, researchers need to collect a large number of laboratory indicators, only a few of which are correlated with mortality or hospitalization rate \cite{yan2020interpretable}. For the MIMIC dataset \cite{johnson2016mimic,johnson2023mimic}, researchers need to ignore some of the patient data (e.g., Doctors record patient conditions using natural language) during the data input stage \cite{jiang2024reasoning, ma2024clibench, chen2024clinicalbench, yu2025intellicare, fang2025toward,yan2024clinicallab}. However, there may still be a lot of redundant information. Because the patient's EHR still contains a lot of laboratory indicators, disease diagnosis records, and medication records, etc. 

In addition to using real clinical data, redundant information can also be added to the clinical context manually. Some work directly inserts irrelevant information into the clinical context as part of red team attacks \cite{pan2025beyond, balazadeh2025red}. There are also some works showing that redundant information has an impact on LLM reasoning performance \cite{rajeev2025cats}. 

\subsection{Incomplete Information Background}
The clinical background information can be incomplete for doctors or clinical decision makers. That is, doctors can only obtain part of the clinical information initially. Some earlier works studied the ability of LLMs to diagnose multi-turn dialogues, and they often used online doctor-patient dialogue for evaluation \cite{xu2024data, zhang2023huatuogpt, bao2023disc, qiu2023smile}. Online consultation is a new medical method, but it may deviate from the real clinical environment. In order to simulate the real clinical environment, the existing task set an incomplete information background as a multi-agent model. These tasks at least include two agents: the doctor agent and the patient agent. The doctor agent needs to communicate with the patient agent, ask the patient agent to perform the corresponding laboratory tests, gradually obtain clinical information, and finally make clinical decisions. The doctor agent is responsible for asking questions, and the patient agent retrieves the corresponding information from the dataset to answer the doctor agent's questions. Researchers often use existing datasets to create benchmarks. The datasets include MedQA \cite{schmidgall2024agentclinic, johri2025evaluation, li2024mediq}, PMC-OA \cite{qiu2025quantifying}, MIMIC \cite{schmidgall2024agentclinic, hager2024evaluation} and so on.

\subsection{Summary}
We categorize the clinical background settings into four types. To our knowledge, the first two settings are often found in medical qualification exams, while the latter two settings are closer to clinical settings. Some datasets and benchmarks may have more than one clinical background setting. For example, a benchmark that uses both the MedQA and MIMIC datasets may have two clinical background settings \cite{liu-etal-2024-large}. Existing papers all show that the performance of the same dataset in an incomplete information background setting is significantly lower than that in a complete information background setting \cite{li2024mediq}. 

\section{Clinical Questions}
Clinical questions are part of clinical decision-making. In this section, we introduce various question settings for clinical decision-making.

\subsection{True/False and Multiple Choice Questions.} 
True/False and multiple-choice questions are collectively called closed-end questions; they are also the question formats used in most medical datasets and benchmarks \cite{pal2022medmcqa, jin2021disease, liu2023benchmarking,hendrycks2020measuring}. Some questions have infinite solution spaces, such as the problem of hospitalization days prediction, but some existing work stipulates it to a specific range and turns it into a multiple-choice question \cite{chen2024clinicalbench, yu2025intellicare, fang2025toward}. The advantage of multiple-choice questions is that the correctness of the answers is relatively easy to judge, and it is straightforward to design evaluation metrics to calculate the performance of different LLMs. However, their decision-making process cannot be effectively explained. The number of options in a multiple-choice question significantly affects the evaluation of model performance. If a multiple-choice question has only 4 options, then even if the model randomly selects, it will have a 25\% accuracy rate. Some work recognizes the issues and therefore expands the number of options to 10 to reduce the accuracy of random selection \cite{wang2024mmlu, zuo2025medxpertqa}. This makes the multiple-choice questions closer to the short-answer question setting and clinical environment.

\subsection{Short Answer Questions}
Compared to multiple-choice questions, short-answer questions have a larger or even infinite solution space. Sometimes, this type of question is also considered a special type of closed-ended question. For example, some questions require output of disease names or disease codes \cite{zhu2025diagnosisarena, ma2024clibench, soroush2024large, xu2025medagentgym}. The current ICD system has about 70,000 disease codes. Some questions need to output a numerical value, such as the dosage of a patient's medication \cite{khandekar2024medcalc}. Compared to open-ended questions, their correctness is easier to verify. For example, if we give a medication dosage, we can easily compare it with the groundtruth to determine whether it is the correct answer. In open-ended questions, sentences and paragraphs of different styles may be the correct answer. 

\subsection{Open-Ended Questions}
An open-ended question is a question that cannot be answered with a "yes" or "no" response or with a static response. Open-ended questions are phrased as a statement that requires a longer answer. Different styles of answers or statements can all be correct in open-ended questions. Open questions can be knowledge-based questions \cite{singhalLargeLanguageModels2023}. They can also be questions involving complex clinical scenarios, such as medical ethics issues \cite{balas2024exploring}. In addition, open-ended questions can often be generated directly from closed-ended questions, such as removing the options of closed-ended questions \cite{sandeep2024few}. Some generative tasks, such as summarizing electronic health records, can also be viewed as open-ended questions. 

\subsection{Summary} 

We categorize the clinical question settings into four major types. To our knowledge, the first two settings are often found in medical qualification exams, while the latter two settings are closer to clinical settings. Figure 3 shows the clinical background and clinical settings used by existing mainstream datasets/benchmarks.

\begin{figure*}[!t]
\centering
\includegraphics[width=2\columnwidth]{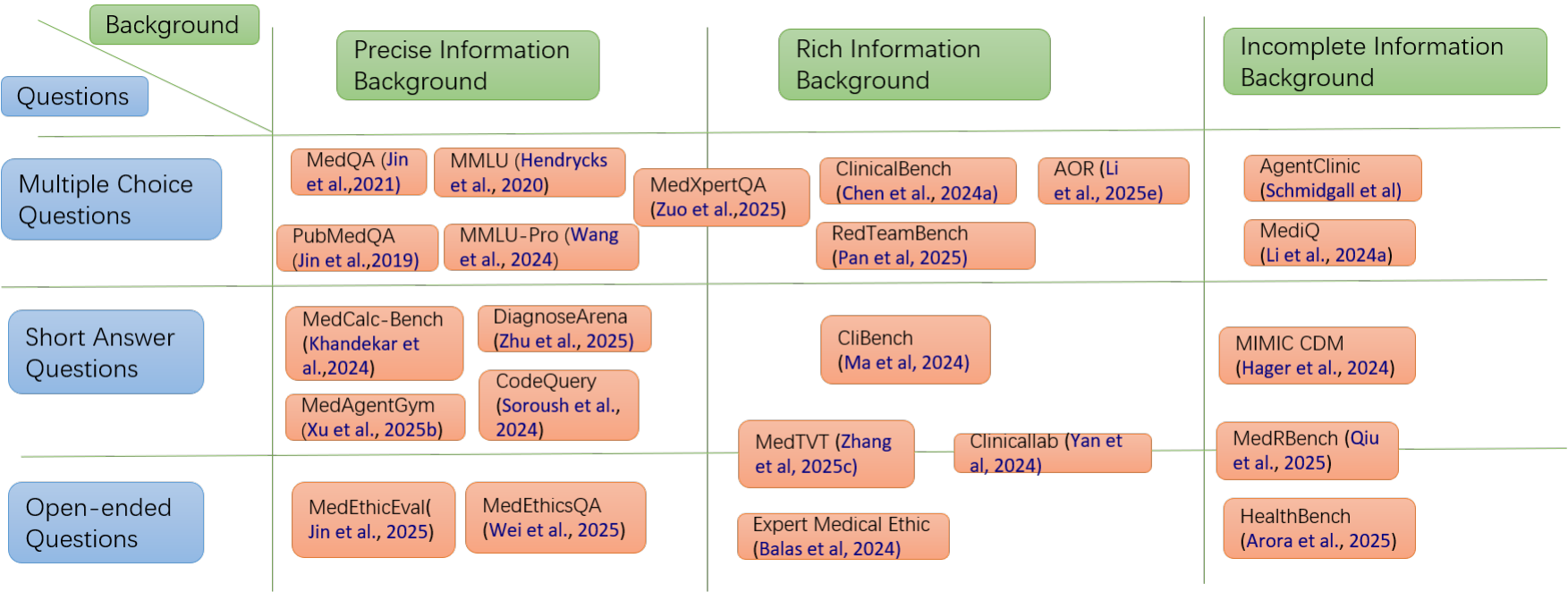} 
\caption{The clinical background and clinical question settings of mainstream medical datasets/benchmarks. Note that some datasets/benchmarks may have multiple settings.}
\label{basic_factors}
\end{figure*}

\section{Methods for Clinical Decision-Making}
In this section, we will introduce methods used in clinical decision-making. Although most of these methods were applied to clinical settings with precise information and multiple-choice questions, they still have implications for real-world clinical decision-making. We can categorize them into two types: training-time techniques and test-time techniques. Training-time techniques fundamentally change the model weights to achieve better reasoning capabilities. Test-time techniques guide and optimize the model’s output during reasoning without modifying the model itself. 

\subsection{Training-time Techniques}
In this paper, we divide training-time techniques into two categories: supervised fine-tuning (SFT) and reinforcement learning (RL). Although in some works, reinforcement learning is also considered a form of supervised fine-tuning \cite{dong2023abilities}. We believe that the primary difference between them is that SFT utilizes explicit groundtruth as a supervisory signal, whereas reinforcement learning can employ "soft constraints" such as reward functions and human preferences, and does not require groundtruth mandatorily \cite{wang2025medical}. 

\textbf{Supervised fine-tuning (SFT)}. Recent work has explored strategies for fine-tuning large language models in multiple stages using high-quality synthetic data \cite{yu2025finemedlm}. There are also some works exploring strategies for supervised fine-tuning in multimodal backgrounds, which is one of the rich Information backgrounds \cite{li2025aor, chen2024miss}. In the context of incomplete information (generally multi-round doctor-patient dialogue), supervised fine-tuning using answers to multiple-choice questions or short-answer questions as supervisory signals has also begun to emerge \cite{li2025aligning, ding2025promed}. These works have shown the effectiveness of SFT in incomplete information backgrounds and multiple-choice questions.

\textbf{Reinforcement learning (RL)}. 
In the development of large language models, many reinforcement learning methods have emerged, such as proximal policy optimization (PPO) \cite{schulman2017proximal}, direct preference optimization (DPO) \cite{rafailov2023direct}, and group relative policy optimization (GRPO) \cite{shao2024deepseekmath}. Currently, DPO and GPRO have begun to be applied to clinical environment settings due to their advantages, such as being lightweight and having no need for value networks. For example, GPRO is used for optimization and alignment in multimodal clinical backgrounds \cite{xu2025medground}, and DPO is used for optimization and alignment in multi-round dialogues \cite{li2025aligning, ding2025promed}.

\subsection{Test-time Techniques}
Common test-time techniques include Chain-of-Thought (CoT) \cite{wei2022chain}, retrieval-augmented generation (RAG) \cite{lewis2020retrieval}, multi-agent systems, etc. 

\textbf{Chain-of-Thought (CoT)}. CoT facilitates problem-solving by guiding the model through a step-by-step reasoning process by using a coherent series of logical steps. There are many ways to decompose a medical task into different steps. For example, in clinical diagnostic tasks, some works decompose the task into the doctor's underlying reasoning process \cite{wu2023large, chen2024cod, zhou2025pathcot}. In the medical dialogue generation task, the task is decomposed into simple subtasks \cite{he2024bp4er}.

\textbf{Retrieval-augmented generation (RAG)}. RAG is a technique that enhances LLM by retrieving relevant information from external knowledge sources before generating a response, making answers more accurate, timely, and grounded in verifiable facts. Early work directly searched databases such as PubMed or Wikipedia \cite{xiong2024benchmarking}. But now, some works have first constructed these databases into knowledge graphs, which greatly improves efficiency when searching in LLM \cite{wu2025medreason, wu2024medical, wen2023mindmap}.

\textbf{Multi-agent systems}. Multi-agent systems consist of multiple intelligent, autonomous agents that collaborate, coordinate, and interact in a shared environment to solve complex problems. There are many works that use multi-agent collaboration to solve medical problems \cite{kim2024mdagents, ke2024enhancing, zhang2025multi}. In addition, multi-agent systems have unique advantages in simulating doctor-patient interactions. Using doctor agents and patient agents can simulate real medical scenarios. Currently, there are many works on simulating real medical scenarios based on multi-agent systems, ultimately improving the diagnostic capabilities of doctor agents after doctor-patient interactions \cite{fan2024ai, li2024agent}

\subsection{Summary}
We have introduced a variety of training-time and test-time techniques. However, these techniques are not mutually exclusive and can be used in combination. For example, in multi-stage learning, SFT can be used in the first stage and RL in the second stage \cite{ding2025promed}. We can also use multi-agent systems, RAG, and CoT at the same time during the testing phase.

\section{Evaluation and Benchmarking}
In this section, we introduce metrics for evaluating the performance of LLMs in clinical decision-making tasks. 

\subsection{Effectiveness}
For closed-ended questions, accuracy is an easy-to-use metric to measure effectiveness, which is calculated as follows:

\begin{equation}
Accuracy = a_c/a_t
\end{equation}

$a_c$ is the number of samples predicted correctly, $a_t$ is the number of all samples. It reflects the match or multiple-choice score on known-answer questions. On some datasets, such as MedQA, LLMs such as Med-PaLM2 \cite{singhal2025toward} have reached or exceeded the expert level. However, the clinical background and answer settings of these datasets are relatively simple. Recent datasets have increased the length of text in clinical contexts, added data from other modalities such as images and time series data, and increased the number of multiple-choice questions. These changes have made multiple-choice questions more complex, and even the most advanced model GPT-4o cannot achieve an accuracy of 50\% \cite{zuo2025medxpertqa}. 

\subsection{Efficiency} 
Efficiency is crucial for complete and redundant information backgrounds and incomplete information backgrounds. Doctors or clinical decision makers must extract key information from complex and chaotic information; ask patients accurate questions, and require patients to undergo appropriate examinations to ensure the timeliness and effectiveness of clinical decisions. Some recent studies have explored methods for evaluating efficiency. They mainly use the LLM-as-a-judge method to evaluate the proportion of key questions or key dialogues in the doctor-patient dialogue \cite{qiu2025quantifying}. Specifically, this work uses high-performance LLM (such as GPT-4o) to divide the reasoning step into $N$ sub-steps: ${r_1,r_2,...r_N}$, For each step, a binary variable $e_i$ is used to indicate whether it is on the correct reasoning path. $e_i = 1$ means that the step refers to steps that provide additional insights and contribute to the final decision, and does not repeat the previous steps; otherwise, it is $0$. Efficiency is calculated according to the following formula:

\begin{equation}
Efficiency = \frac{1}{N}\sum_{i=1}^{N}e_i
\end{equation}

\subsection{Explainability} 
In medical decisions, the decision maker or patient needs to give a reasonable explanation for the answer rather than simply giving an answer. This is especially important in short-answer questions and open-ended questions, especially open-ended questions. There are currently two main ways to evaluate LLM explainability: manual evaluation and LLM-as-a-judge. 

Manual evaluation often develops a series of scoring rules based on the task, such as using a Likert scale \cite{balas2024exploring}, and finally averages the scores of multiple people to get the final score \cite{jin2025medethiceval}. In some studies, such as \emph{Healthbench} \cite{arora2025healthbench}, these scoring settings are more complex, requiring not only the accuracy of the answers, but also the order in which different contents appear.

LLM-as-a-judge generally uses LLMs with stronger reasoning ability for scoring. It can be expressed by the following formula \cite{gu2024survey}:

\begin{equation}
\mathcal{E} = \mathcal{P_{LLM}}(x\oplus\mathcal{C})
\end{equation}

$\mathcal{E}$ is the final evaluation obtained from the whole LLM-as-a-Judge process, it can be scores or other metrics. $\mathcal{P_{LLM}}$ is the large language model that use to evaluate. $x$ is the data which waiting to be evaluated, it is generally the output of a large language model in a clinical case. $\mathcal{C}$ The context for the input $x$, which is often prompt template. Essentially, it is a replacement for human evaluation. Therefore, some of the rules used in human evaluation can be used as context for LLM-as-a-judge; $\mathcal{C}$ is often the scoring criterion specified by the user.
A common evaluation method is the reasoning process is broken down into several parts and scored based on whether the answer matches each part \cite{qiu2025quantifying, wei2025medethicsqa}. For example, if the answer is divided into 4 steps and the full score of each step is 1 point, when LLM evaluates the answer, it is considered that 2 steps have 1 point, 1 step has 0.5 points, and 1 step has no points, so the final score is 2.5 points.

\section{Open Issues}

In this section, we summarize some of the current status in clinical decision-making research and raise some open issues.

\subsection{Creating Datasets and Benchmarks relevant to Real Clinical Background} 
At present, most of the work is still limited to the clinical environment with precise information backgrounds and multiple-choice questions. Although the background settings are becoming more and more complex and the multiple-choice options are increasing, they still do not break out of this fundamental framework \cite{zuo2025medxpertqa}. An important leap from a complete information context to a real clinical context with redundant information is that most redundant information is real laboratory test data, which is difficult to synthesize artificially. It is necessary to create a multi-task clinical benchmark full of open-ended questions.

\subsection{Filtering Redundant Information} 
The real clinical environment has to deal with a lot of redundant information. Long contexts already pose a challenge to the reasoning of LLMs \cite{liu2025comprehensive}, and redundant information may further degrade the reasoning performance of LLMs \cite{rajeev2025cats}. Therefore, it is crucial to guide LLMs to filter out redundant information. In the context of incomplete information, this question can be further extended to: how to get the doctor's agent to ask a "good" question \cite{li2025aligning,fang2025toward}, this means that doctors need to ask questions that are closely related to the illness the patient may be diagnosed with. 

\subsection{Multi-agent systems suitable for clinical environment} 
Multi-agent systems have been used in medical question answering and have proven effective. In data-rich clinical environments, however, multi-agent systems offer unique advantages. This is because the context in clinical settings is often longer. By splitting clinical tasks into subtasks, each processing only a portion of the data, multi-agent systems can avoid the issues caused by the long context of a single agent.

In addition, multi-agent systems are a good way to simulate real clinical environments. Multi-agent dialogue is crucial for simulating the doctor-patient interaction process in an incomplete information environment. There is currently little research on patient agents in doctor-patient dialogues. Some work has studied the fairness, trust, and bias issues of patient agents \cite{schmidgall2024agentclinic}, but there is insufficient research on whether patient agents can effectively retrieve datasets and accurately answer doctors' questions. 

\subsection{Designing New Training Paradigms for Open-ended Questions} 
In "Closed-end questions" such as multiple-choice questions and short-answer questions, whether it is supervised fine-tuning or reinforcement learning, the existence of a standardized groundtruth makes it easier to set the loss function/reward. However, in open-ended questions, there may be multiple answers that meet the requirements, and there is no standard groundtruth for the answers. How to design the loss function or reward is a problem that needs to be solved. Some label-free reinforcement learning methods may be the key to solving this problem \cite{zhou2025evolving}. 

\subsection{New Modeling Strategies for Evaluation}
Currently, there are few evaluation metrics for efficiency, and there are no evaluation efficiency metrics specifically designed for multi-round doctor-patient dialogues. In terms of explainability, human evaluation is too expensive to evaluate clinical open-ended questions, while using LLM-as-a-judge faces issues of reliability, robustness, and hallucinations \cite{gu2024survey}. Design efficiency evaluation metrics and design a better LLM-as-a-judge method to evaluate clinical questions, especially open-ended questions, as a possible future research direction.

\section{Conclusion}
This review analyzes the current status and challenges of LLMs in clinical decision-making. We first divide clinical decision-making tasks based on the two dimensions: clinical background and clinical questions. Then, we further explore the current status of methods and evaluation metrics from these two dimensions. Finally, we summarize the challenges faced in clinical decision-making.

\section{Limitations}
Our study still has some limitations. First, our study may have overlooked some recent preprints. Second, our study provides a classification perspective on clinical decision-making. However, there may still be other classification perspectives on clinical decision-making. For example, clinical decision-making can be classified by task type.

\bibliography{custom}

\appendix

\end{document}